\documentclass{article}
\usepackage{ijcai22}

\usepackage{times}
\usepackage{soul}
\usepackage{url}
\usepackage[hidelinks]{hyperref}
\usepackage[utf8]{inputenc}
\usepackage[small]{caption}
\usepackage{graphicx}
\usepackage{amsmath}
\usepackage{amsthm}
\usepackage{booktabs}
\usepackage[ruled]{algorithm2e}
\usepackage{graphicx}

\usepackage{tikz}
\usepackage{comment}
\usepackage{amsmath,amssymb} 
\usepackage{color}
\usepackage{xcolor}

\usepackage{graphicx}
\usepackage{amsmath}
\usepackage{amssymb}
\usepackage{booktabs}
\usepackage{amsmath}
\usepackage{tabularx}

\title{Vision Big Bird: Random Sparsification for Full Attention}

\author{
Zhemin Zhang$^1$
\and
Xun Gong$^1$
\affiliations
$^1$Southwest Jiaotong University
\emails
zheminzhang@my.swjtu.edu.cn
}

\begin{document}

\maketitle

\begin{abstract}
Recently, Transformers have shown promising performance in various vision tasks. However, the high costs of global self-attention remain challenging for Transformers, especially for high-resolution vision tasks. Inspired by one of the most successful transformers-based models for NLP: Big Bird, we propose a novel sparse attention mechanism for Vision Transformers (ViT). Specifically, we separate the heads into three groups, the first group used convolutional neural network (CNN) to extract local features and provide positional information for the model, the second group used Random Sampling Windows (RS-Win) for sparse self-attention calculation, and the third group reduces the resolution of the keys and values by average pooling for global attention. Based on these components, ViT maintains the sparsity of self-attention while maintaining the merits of Big Bird (i.e., the model is a universal approximator of sequence functions and is Turing complete). Moreover, our results show that the positional encoding, a crucial component in ViTs, can be safely removed in our model. Experiments show that Vision Big Bird demonstrates competitive performance on common vision tasks.
\end{abstract}

\section{Introduction}

Modeling in computer vision has long been dominated by convolutional neural networks (CNNs). Transformers \cite{NIPS2017-3f5ee243}, originated from natural language processing (NLP), have recently demonstrated state-of-the-art performance in visual learning. Despite tremendous successes, the full self-attention have computational and memory requirement that is quadratic in the sequence length, which, in turn, greatly limits its applicability to high-resolution vision tasks. To reduce the computation cost, numerous efforts have studied how to introduce the locality of the CNN model into the ViT to improve its scalability. For example, Swin Transformer \cite{Liu-2021-ICCV} designs window attention, and MaxViT \cite{10.1007/978-3-031-20053-3-27} designs grid attention. However, these methods \cite{Liu-2021-ICCV,Dong-2022-CVPR,Vaswani-2021-CVPR} use window-based attention at shallow layers, losing the non-locality of original ViT, which results in limited model capacity and henceforth scales unfavorably on larger datasets such as ImageNet-21K \cite{NEURIPS2021-20568692}. To bridge the connection between windows, previous works propose specialized designs such as the haloing operation \cite{Vaswani-2021-CVPR} and shifted window \cite{Liu-2021-ICCV}. These approaches often need complex architectures, and their receptive field is increased quite slowly and requires stacking many blocks to achieve global self-attention.

\begin{figure}[t]
  \centering
   \includegraphics[width=1.0\linewidth]{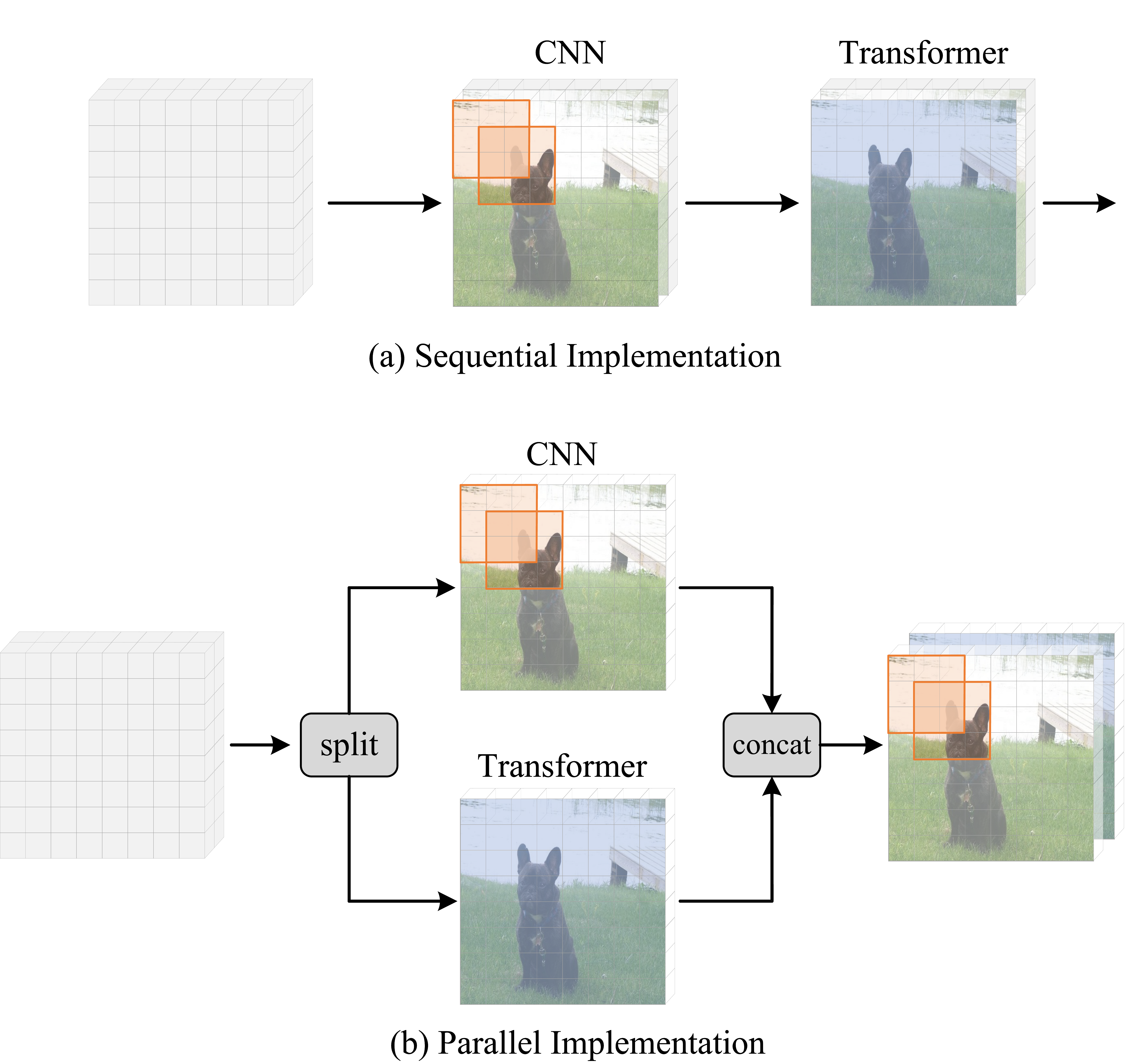}
   \caption{Parallel vs. Sequential. (a) Previous works usually introduce CNNs in a sequential manner. (b) In Vision Big Bird, we introduce CNNs in parallel, so that each block contains both global and local information. When we replace the Parallel manner with Sequential, the performance of Vision Big Bird degrades on all tasks.}
   \label{ParallelandSequentialCNN-flabel}
\end{figure}

The Big Bird model \cite{NEURIPS2020-c8512d14} is a Transformer for modeling longer sequences and is primarily built on top of ETC \cite{DBLP:journals/corr/abs-2004-08483}. The Big Bird model does not introduce new parameters beyond the Transformer model and the memory complexity of the self-attention is linear, i.e., O(n). Importantly, Big Bird is shown to be a universal approximator of sequence functions and is Turing complete. The form of Big Bird can be summarized as: \textit{windowed attention + random attention + global attention}. Inspired by Big Bird, we propose a novel sparse attention mechanism for Vision Transformers that reduces the computation cost while maintaining the merits of Big Bird (i.e., the model is a universal approximator of sequence functions and is Turing complete). The form of Vision Big Bird (VBB) can be summarized as: \textit{convolution+ random sampling windows + global attention}.

\begin{figure}[t]
  \centering
   \includegraphics[width=0.8\linewidth]{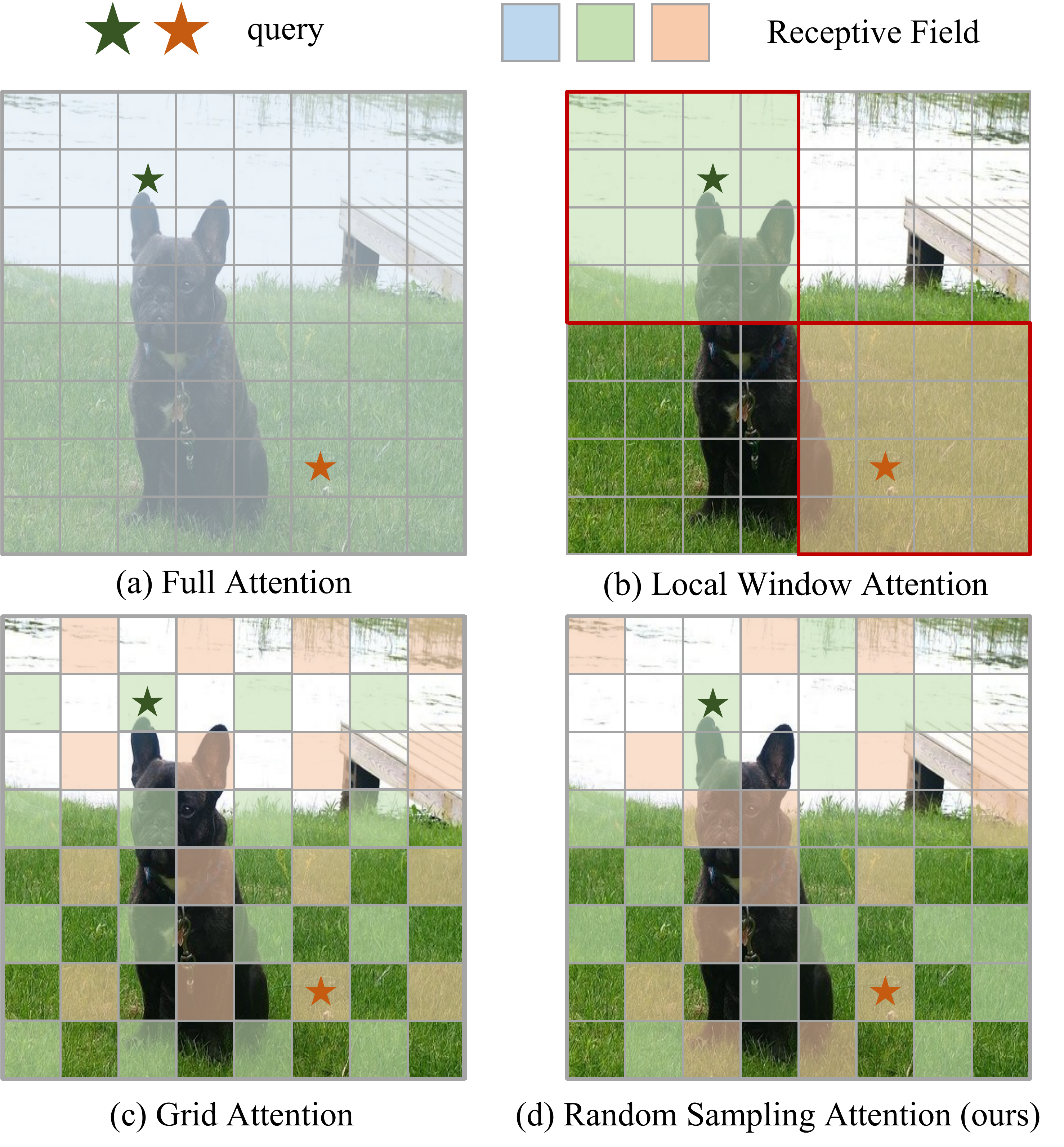}
   \caption{ Full attention and its sparse variants. (a) Full attention is a global operation that is computationally expensive and requires a lot of memory. (b) Swin Transformer uses partitioned window attention. (c) Grid attention attends globally to patches in a sparse, uniform grid overlaid on the entire 2D space. (d) Random sampling attention sample random patches compose a window and perform self-attention within the window. The same colors are spatially mixed by the self-attention operation.}
   \label{MultipleAttentionMechanisms-flabel}
\end{figure}

The original ViT lacks inductive bias, such as locality and translation equivariance, which leads to overfitting and data inefficient usage. To improve data efficiency, recent work has proposed a series of hybrid Vision Transformers that combine CNNs with Vision Transformers \cite{Ren-2023-ICCV,10.1007/978-3-031-20053-3-27,Wu-2021-ICCV}. In VBB, instead of introducing CNNs in a sequential manner as in previous work, we introduce CNNs in parallel, as shown in Figure \ref{ParallelandSequentialCNN-flabel}. By utilizing the local receptive fields, shared weights, and spatial subsampling of CNNs, VBB can capture the local information of images, and thus also achieves some degree of shift, scale, and distortion invariance. By introducing CNNs in parallel, the locality of CNNs is fused in every block of VBB, and in our experiments, the performance of parallel introduction of CNNs is much higher than sequential introduction. Meanwhile, since CNNs can implicitly learn the positional information, our results show that the positional encoding, a crucial component in Vision Transformers, can be safely removed in VBB.

Taking inspiration from graph sparsification methods, we propose a novel Transformer module called random sampling windows (RS-Win). RS-Win sample random image patches to compose the window, following a uniform distribution, i.e., the patches in RS-Win can come from any position in the image, shown in Figure \ref{MultipleAttentionMechanisms-flabel} (d). RS-Win can perform both local and global spatial interactions in a single block. Compared to the previously handcrafted sparse attention, RS-Win gives greater flexibility. The number of patches randomly sampled in each window of RS-Win is fixed, and computing self-attention locally within windows, thus the complexity becomes linear to image size.

Now we need to obtain global information as well. Since adding additional global tokens breaks the regular structure of structured data (e.g., images), this will result in the Vision Transformer cannot build hierarchies efficiently. Therefore, unlike Big Bird, which uses global tokens to capture global information, in VBB we maintain the resolution of queries and use average pooling to reduce the resolution of keys and values. By controlling the granularity of pooling at different stages, the computation of global attention at each stage is linear.

Finally, we connect the features from the output of these three modes and feed the results into subsequent MLP layers for integration. Based on the above components, we design a general Vision Transformer backbone with a hierarchical architecture, named Vision Big Bird (VBB). VBB benefits from local, random and global receptive fields throughout the entire network, from shallow to deep stages, demonstrating superior performance in regards to both model capacity and generalization abilities. Our tiny variant VBB-T achieves 83.8\% Top-1 accuracy on ImageNet-1K without any extra training data.

\begin{figure*}[t]
\centering
\includegraphics[width=0.9\linewidth]{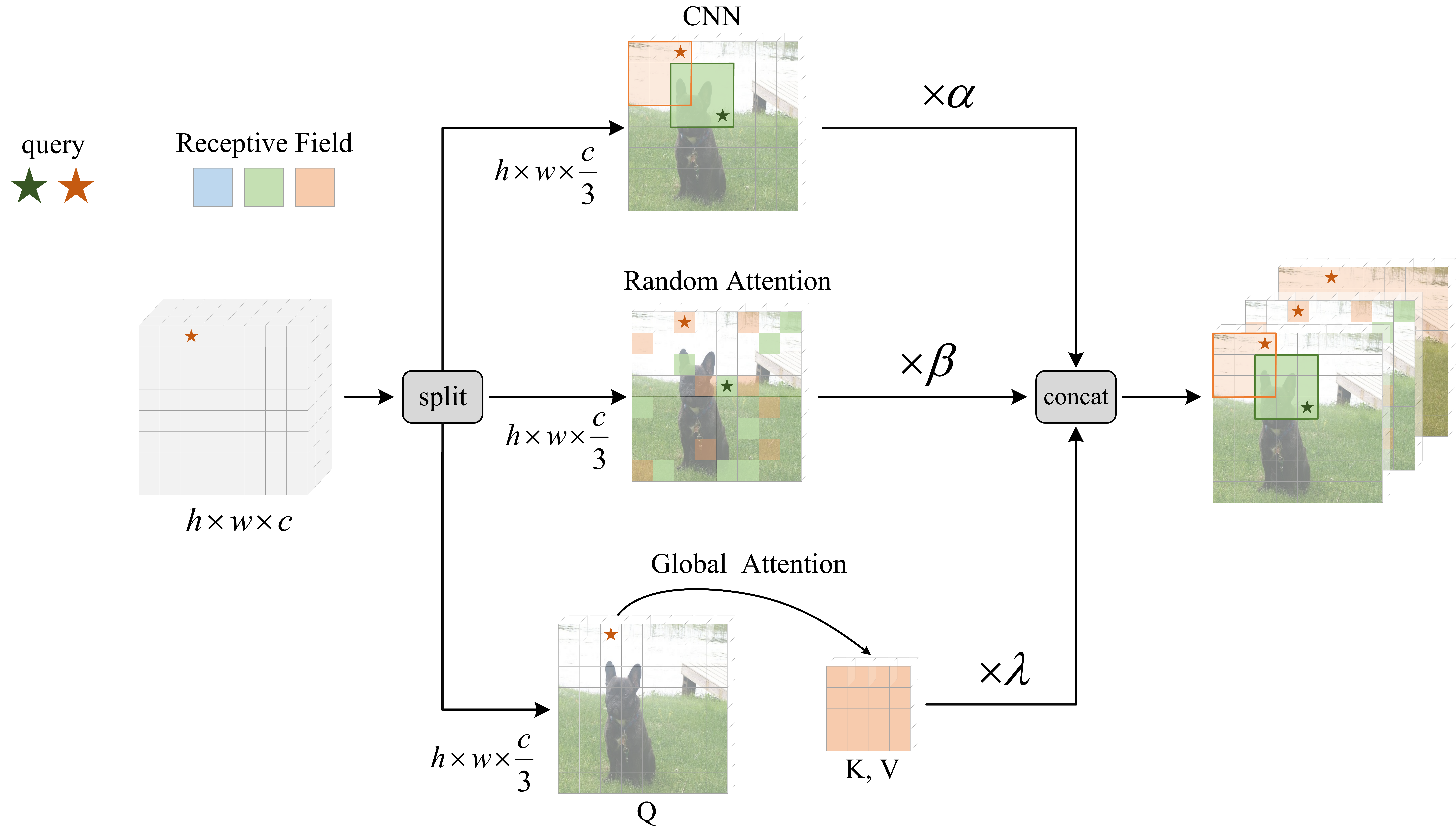} 
\caption{The overall architecture of Vision Big Bird. We split the multi-heads of self-attention into three parallel groups for Convolution, RS-Win and Global Attention, respectively. Each mechanism is weighted using $\alpha$,$\beta$,$\lambda$ before the output is connected, allowing the model to adaptively learn the importance of different mechanisms.}
\label{OverallArchitecture-flabel}
\end{figure*}

\section{Related Work}

Transformers were proposed by Vaswani et al. \cite{NIPS2017-3f5ee243} for machine translation, and have since become the state-of-the-art method in many NLP tasks. Big Bird \cite{NEURIPS2020-c8512d14} is one of the most successful transformers-based models, proposing a sparse attention mechanism that reduces this quadratic dependency of the full attention to linear. Recently, ViT \cite{DBLP:journals/corr/abs-2010-11929} demonstrates that pure Transformer-based architectures can also achieve very competitive results.  One challenge for Vision Transformer-based models is data efficiency. Although ViT can perform better than convolutional networks with hundreds of millions of images for pre-training, such a data requirement is difficult to meet in many cases.

To improve data efficiency, many recent works have focused on introducing the locality and hierarchical structure into ViT. The Swin Transformer \cite{Liu-2021-ICCV} pays attention on shifted windows in a hierarchical architecture.  Nested ViT \cite{zhang2022nested} proposes a block aggregation module, which can more easily achieve cross-block non-local information communication. HiLo Transformer \cite{NEURIPS2022-5d5f703e} disentangles the high/low frequency patterns in an attention layer by separating the heads into two groups, where the high frequencies capture local fine details and low frequencies focus on global structures. Based on the local window, a series of local self-attentions with different shapes are proposed in subsequent work. Axial self-attention \cite{DBLP:journals/corr/abs-1912-12180} and Criss-cross attention \cite{Huang-2019-ICCV} achieve longer-range dependencies in horizontal and vertical directions respectively by performing self-attention in each single row or column of the feature map. CSWin \cite{Dong-2022-CVPR} proposed a cross-shaped window self-attention region, including multiple rows and columns. The performance of the above attention mechanisms are either limited by the restricted window size or has a high computation cost, which cannot achieve a better trade-off between computation cost and global-local interaction.

On the other hand, many works have blended attention model with convolution, and accordingly propose a series of hybrid Vision Transformers. CvT \cite{Wu-2021-ICCV} designs a hierarchy of Transformers containing a new convolutional token embedding, and a convolutional Transformer block leveraging a convolutional projection, introducing desirable properties of CNNs to the ViT architecture. MaxViT \cite{10.1007/978-3-031-20053-3-27} proposed grid attention and designs a new architectural element that effectively blending grid attention with convolutions, where the grid attention module attends globally to pixels in a sparse, uniform grid overlaid on the entire 2D space. FasterViT \cite{hatamizadeh2023fastervit} combines the benefits of fast local representation learning in CNNs and global modeling properties in ViT, and the proposed Hierarchical Attention (HAT) approach decomposes global self-attention with quadratic complexity into a multi-level attention with reduced computational costs. 

\begin{figure}[h]
  \centering
   \includegraphics[width=0.4\linewidth]{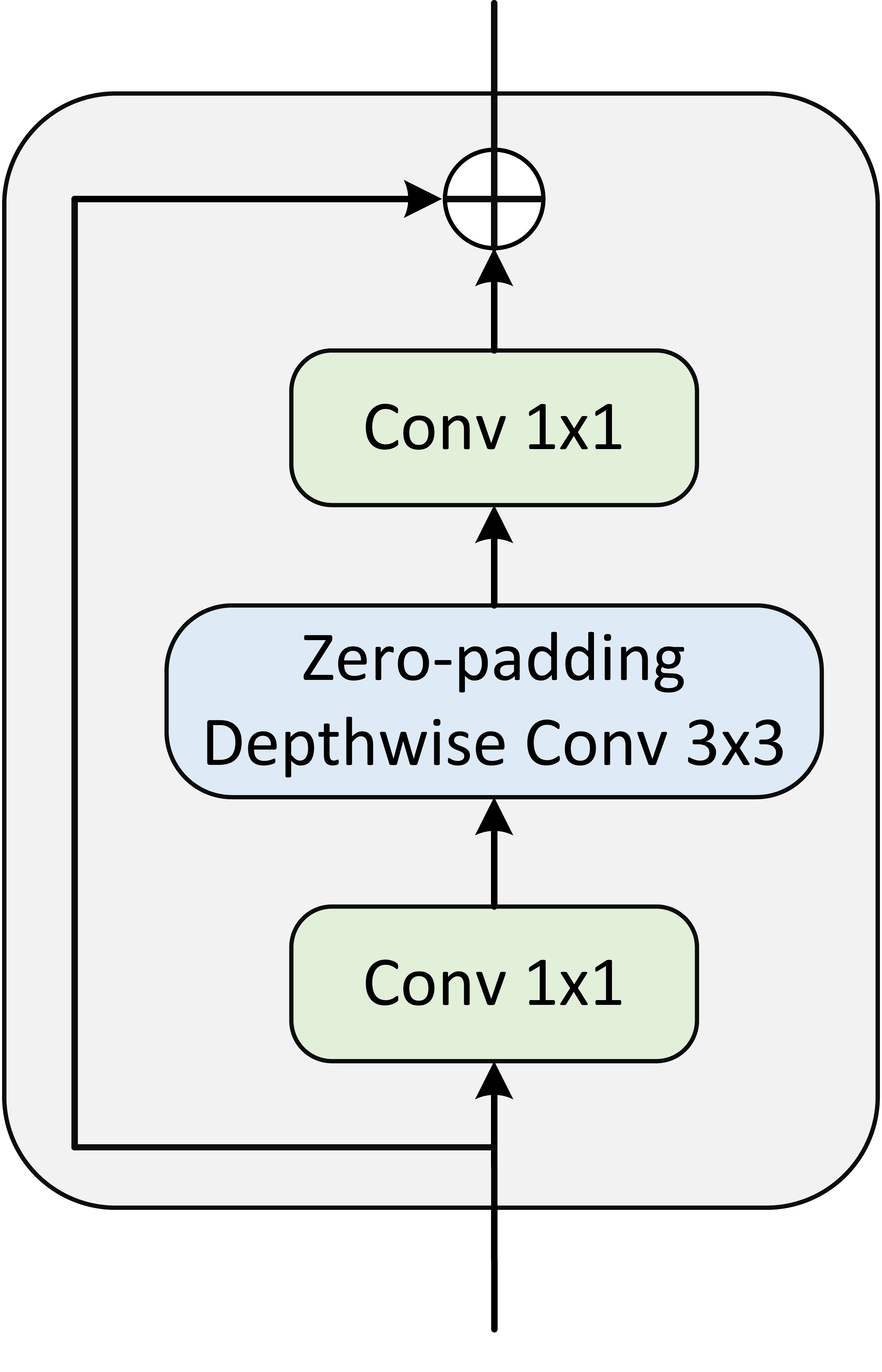}
   \caption{Convolutional block with zero-padding.}
   \label{CNNLayer-flabel}
\end{figure}

\section{Method}

Inspired by the sparse approaches presented in Big Bird \cite{NEURIPS2020-c8512d14}, we introduce a new type of attention module, named Vision Big Bird, as shown in Figure \ref{OverallArchitecture-flabel}. The VBB decomposes the fully dense attention mechanisms into three sparse forms: convolution, random attention, and global attention, which reduces the quadratic complexity of vanilla attention to linear, without any loss of non-locality. Essentially, VBB can be simply viewed as a mechanism to reduce computation through random sparsification while supplementing local and positional information with CNNs and global information with global attention. In the next, we describe the three mechanisms in detail.

\subsection{Zero-padded convolutional layers}

In the convolutional branch of the VBB we follow a typical hierarchical design of CNNs (e.g., EfficientNet \cite{pmlr-v139-tan21a}), but use a new type of convolutional layer. As a recent study \cite{DBLP:journals/corr/abs-2001-08248} has shown that positional information can be implicitly learned from zero-padding in CNNs, we propose to adopt one layer of 3 × 3 depthwise convolutional layer with zero-padding in the convolutional branch to replace the intermediate convolutional layer, as shown in Figure \ref{CNNLayer-flabel}. Normalization and activation layers are omitted for simplicity. By using zero-padding, the output of the convolutional branch of the VBB contains both local and positional information, so positional encoding can be safely removed in the VBB.

\begin{algorithm}[t]
    \caption{\small{ Pseudocode of RS-Win in a PyTorch-like style.}}
    \label{alg:RS-Win}
    \textbf{Input}:{ x.shape=B, L, C (batch, length, dim)}\\
    \textcolor[rgb]{0.133, 0.757, 0.133}{\# sample\_map from uniform distribution}\\
    sample\_map = torch.rand(B, L, device=x.device)\\
    \textcolor[rgb]{0.133, 0.757, 0.133}{\# Sort sample\_map for each token sequence}\\
    ids\_shuffle = torch.argsort(sample\_map, dim=1)\\
    \textcolor[rgb]{0.133, 0.757, 0.133}{\# ids\_restore for sequence restore}\\
    ids\_restore = torch.argsort(ids\_shuffle, dim=1)\\
    \textcolor[rgb]{0.133, 0.757, 0.133}{\# Shuffle token sequence by ids\_shuffle}\\
    x\_shuffe = torch.gather(x, dim=1, index=ids\_shuffle)\\
    \textcolor[rgb]{0.133, 0.757, 0.133}{\#  Partition windows}\\
    x\_windows = window\_partition(x\_shuffe, window\_size)\\
    \textcolor[rgb]{0.133, 0.757, 0.133}{\#  Windows self-attention}\\
    attn\_windows = self.attn(x\_windows)\\
    \textcolor[rgb]{0.133, 0.757, 0.133}{\#  Reverse windows}\\
    x = window\_reverse(attn\_windows, window\_size, H, W)\\
    \textcolor[rgb]{0.133, 0.757, 0.133}{\#  Restore the token sequence}\\
    x\_restore = torch.gather(x, dim=1, index=ids\_restore)\\
\end{algorithm}

\begin{figure}[h]
  \centering
   \includegraphics[width=0.8\linewidth]{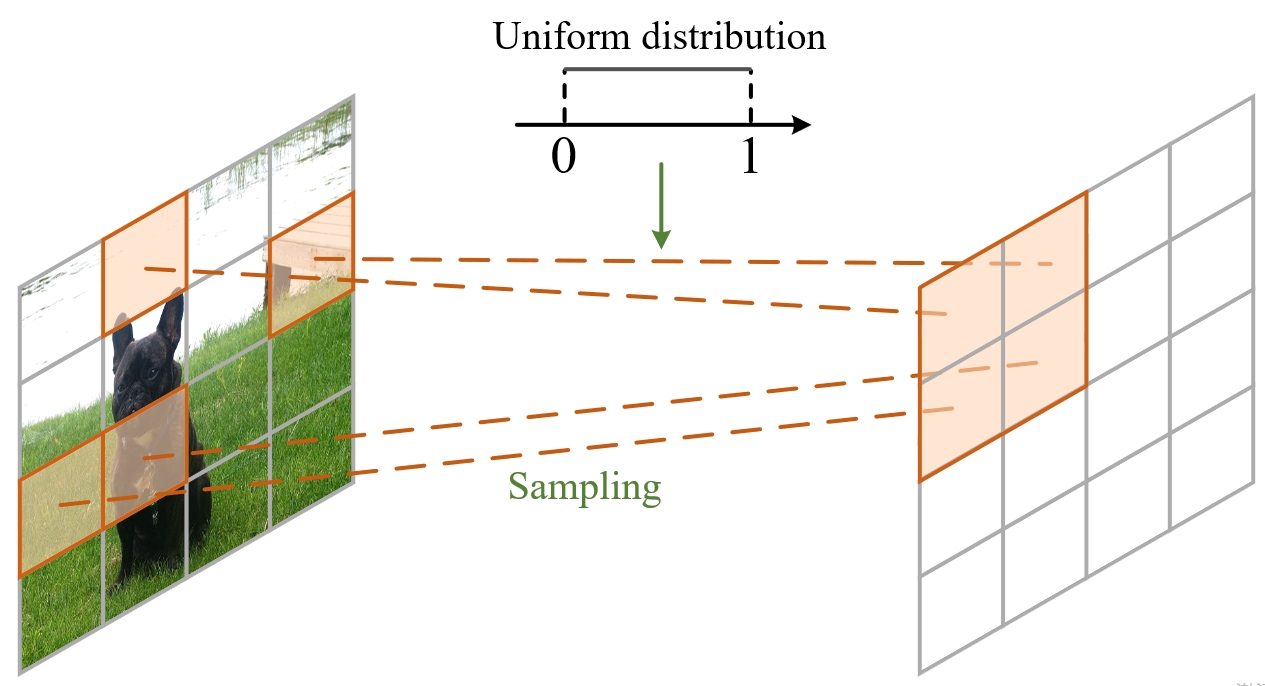}
   \caption{ RS-Win self-attention.}
   \label{RS-Win-flabel}
\end{figure}

\subsection{Random Sampling Windows}

Local ViT uses window-based attention at shallow layers, lacking the non-locality of original ViT, which leads to Local ViT having limited model capacity and henceforth scales unfavorably on larger datasets. Existing works use specialized designs, such as the shifted window \cite{Liu-2021-ICCV}, to communicate information between windows. These approaches often need complex architectures, and their receptive field is increased quite slowly and requires stacking many blocks to achieve global self-attention. For capturing dependencies varied from short-range to long-range, we propose RS-Win self-attention. Compared to the previously handcrafted sparse attention, RS-Win gives greater flexibility.

RS-Win sample random image patches to compose the window, following a uniform distribution, i.e., the patches in RS-Win can come from any position in the image, shown in Figure \ref{RS-Win-flabel}. The RS-Win algorithm is summarized with Pytorch-like pseudo code in Algorithm \ref{alg:RS-Win}. The RS-Win branch of the VBB can be viewed as a random sparsification of full self-attention.

\begin{figure}[h]
  \centering
   \includegraphics[width=1.0\linewidth]{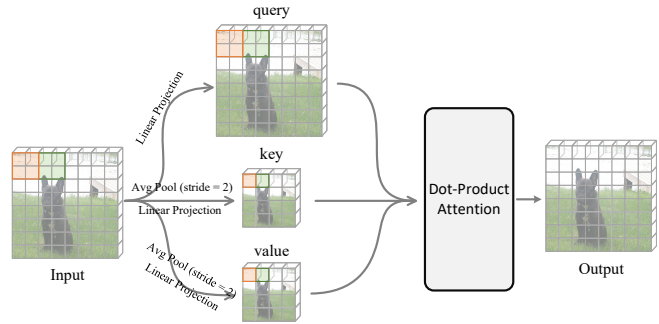}
   \caption{Global self-attention.}
   \label{GlobalAttention-flabel}
\end{figure}

\subsection{Global Attention}

Directly applying multi-head self-attention to high-resolution feature maps requires huge computational cost because full attention is quadratically dependent. Additional added Global Tokens are used in Big Bird to capture global information to avoid the quadratic complexity. However, adding additional Global Tokens breaks the regular structure of structured data (e.g., images), this will result in the Vision Transformer cannot build hierarchies efficiently. Unlike Big Bird, in VBB we maintain the resolution of queries and use average pooling to reduce the resolution of keys and values, shown in Figure \ref{GlobalAttention-flabel}. By controlling the granularity of pooling at different stages, the computation of global attention at each stage is linear.

\subsection{Overall Architecture}

Finally, we merge the above mechanisms together and add different scaling weights for the three mechanisms, where the scaling weights are learnable parameters, as shown in Figure \ref{OverallArchitecture-flabel} ($\alpha$,$\beta$,$\lambda$). By adding scaling weights, the model can adaptively learn the importance of each mechanism, thus improving the model's performance.

Specifically, we split the $K$ heads into three parallel groups, $K/3$ heads per group, thus incorporating three different feature extraction mechanisms. The first group of heads perform CNN, the second group of heads perform RS-Win Attention, and the third group of heads perform Global Attention.

\begin{equation}
\text{hea}{{\text{d}}_{k}}\left\{ \begin{matrix}
   \text{CN}{{\text{N}}_{k}}\left( X \right)  \\
   \text{RS-Wi}{{\text{n}}_{k}}\left( X \right)  \\
   \text{G}{{\text{A}}_{k}}\left( X \right)  \\
\end{matrix} \right.\begin{array}{*{35}{l}}
   k=1,\cdots ,K/3  \\
   k=K/3,\cdots ,2K/3  \\
   k=2K/3,\cdots ,K  \\
\end{array}
\label{splitGroup-glabel}
\end{equation}

Finally, add a scaling weight to each group and concatenate the output of the three parallel groups back together.

\begin{equation}
\text{VBB}\left( X \right)\text{=Concat}\left( \text{CN}{{\text{N}}_{k}}\times \alpha \text{,R}{{\text{S}}_{k}}\times \beta \text{,G}{{\text{A}}_{k}}\times \lambda  \right){{W}^{o}}
  \label{mergingtoken-glabel}
\end{equation}

where ${{W}^{O}}\in {{R}^{C\times C}}$ is the commonly used projection matrix that is used to integrate the output tokens of three groups. $\alpha$,$\beta$,$\lambda$ are learnable scaling weights that allow the model to adaptively set the importance of each mechanism at different stages or in different datasets.

By analyzing the scaling values of the different mechanisms, we find some interesting phenomena. In the ImageNet experiment, the scaling value of Global Attention in the shallow layer of the model is small and increases with the deepening of the layers, while that of RS-Win is opposite. In addition, we find that the scaling values of CNNs in cifar100 experiment is much larger than that in ImageNet, which to some extent can prove the effectiveness of CNN in feature extraction in small datasets. See the subsection 4.3 for more details.

\section{Experiments}

To show the effectiveness of the VBB, we conduct experiments on ImageNet-1K \cite{5206848}. We then compare the performance of VBB and state-of-the-art Transformer backbones on small datasets Cifar-100 \cite{krizhevsky2009learning} and Mini-ImageNet \cite{krizhevsky2012imagenet}. To further demonstrate the effectiveness and generalization of our backbone, we conduct experiments on ADE20K \cite{Zhou-2017-CVPR} for semantic segmentation, and COCO \cite{10.1007/978-3-319-10602-1-48} for object detection. Finally, we perform comprehensive ablation studies to analyze each component of the VBB.

\begin{table}[h]
   \centering
   \caption{Comparison of different models on ImageNet-1K.}
   \resizebox{\linewidth}{!}{
   \begin{tabular}{l|ccc|c}
      \hline 
      Method  & Image Size & Param. & FLOPs                & Top-1 acc.           \\
      \hline 
      RegNetY-4G \cite{Radosavovic-2020-CVPR} & ${{224}^{2}}$   & 21M   & 4.0G     &80.0    \\
      DeiT-S \cite{pmlr-v139-touvron21a} & ${{224}^{2}}$   & 22M   & 4.6G     &79.8    \\
      PVT-S \cite{Wang-2021-ICCV} & ${{224}^{2}}$   & 25M   & 3.8G     &79.8    \\
      Swin-T \cite{Liu-2021-ICCV} & ${{224}^{2}}$   & 29M   & 4.5G     &81.3    \\
      Focal-T \cite{DBLP:journals/corr/abs-2107-00641} & ${{224}^{2}}$   & 29M   & 4.9G     &82.2    \\
      CSWin-T \cite{Dong-2022-CVPR} & ${{224}^{2}}$   & 23M   & 4.3G     &82.7     \\
	 MaxViT-T \cite{10.1007/978-3-031-20053-3-27} & ${{224}^{2}}$   & 31M   & 5.6G     &83.6     \\
      SGFormer-S \cite{Ren-2023-ICCV} & ${{224}^{2}}$   & 22M   & 4.8G     &83.2     \\
      VBB-T (ours) & ${{224}^{2}}$   & 29M   & 5.1G     & \textbf{83.8}     \\
      \hline 
      RegNetY-16G \cite{Radosavovic-2020-CVPR} & ${{224}^{2}}$   & 84M   & 16.0G     &82.9    \\
      ViT-B  \cite{DBLP:journals/corr/abs-2010-11929} & ${{384}^{2}}$   & 86M   & 55.4G     &77.9     \\
      DeiT-B \cite{pmlr-v139-touvron21a} & ${{224}^{2}}$   & 86M   & 17.5G     &81.8    \\
      PVT-B \cite{Wang-2021-ICCV} & ${{224}^{2}}$   & 61M   & 9.8G     &81.7    \\
      Swin-B \cite{Liu-2021-ICCV} & ${{224}^{2}}$   & 88M   & 15.4G     &83.3     \\
      Focal-B \cite{DBLP:journals/corr/abs-2107-00641} & ${{224}^{2}}$   & 90M   & 16.0G     &83.8    \\
      CSWin-B \cite{Dong-2022-CVPR} & ${{224}^{2}}$   & 78M   & 15.0G     &84.2     \\
 	 MaxViT-B \cite{10.1007/978-3-031-20053-3-27} & ${{224}^{2}}$   & 120M   & 23.4G     &84.9     \\
 	 SGFormer-B \cite{Ren-2023-ICCV} & ${{224}^{2}}$   & 78M   & 15.6G     &84.7     \\
      VBB-B (ours) & ${{224}^{2}}$   & 85M   & 15.8G     &\textbf{84.9}     \\
      \hline 
   \end{tabular}
   }
   \label{ImageNet-Top1}
\end{table}

\subsection{Classiﬁcation on the ImageNet-1K}

\noindent \textbf{Implementation details.} This setting mostly follows \cite{Liu-2021-ICCV}. We use the PyTorch toolbox \cite{paszke2019pytorch} to implement all our experiments. We employ an AdamW \cite{kingma2014adam} optimizer for 300 epochs using a cosine decay learning rate scheduler and 20 epochs of linear warm-up. A batch size of 256, an initial learning rate of 0.001, and a weight decay of 0.05 are used. ViT-B/16 uses an image size 384×384 and others use 224×224. We include most of the augmentation and regularization strategies of Swin transformer\cite{Liu-2021-ICCV} in training.

\noindent \textbf{Results.} Table \ref{ImageNet-Top1} compares the performance of the proposed VBB Transformer with the state-of-the-art CNN and Vision Transformer backbones on ImageNet-1K. Compared to ViT-B, the proposed VBB-T model is +5.9\% better and has much lower computation complexity than ViT-B. Compared to other variants of ViT, VBB consistently outperforms competitors under similar parameters and computation costs.  Specifically, VBB achieves 83.8\% and 84.9\% Top-1 Accuracy under tiny and base models, respectively. This compares to 83.6\% and 84.9\% for MaxViT and 83.2\% and 84.7\% for SGFormer.

\begin{table}[h]
   \centering
   \caption{Comparison of different models on Cifar-100.}
   \resizebox{\linewidth}{!}{
   \begin{tabular}{l|ccc|c}
      \hline 
      Method  & Image Size & Param. & FLOPs                & Top-1 acc.           \\
      \hline 
       Swin-T \cite{Liu-2021-ICCV} & ${{224}^{2}}$   & 29M   & 4.5G     &78.8    \\
       Focal-T \cite{DBLP:journals/corr/abs-2107-00641} & ${{224}^{2}}$   & 29M   & 4.9G     &78.5     \\
       CSWin-T \cite{Dong-2022-CVPR} & ${{224}^{2}}$   & 23M   & 4.3G     &81.1     \\
	  MaxViT-T \cite{10.1007/978-3-031-20053-3-27} & ${{224}^{2}}$   & 31M   & 5.6G     &82.6     \\
       SGFormer-S \cite{Ren-2023-ICCV} & ${{224}^{2}}$   & 22M   & 4.8G     &82.3     \\
       VBB-T (ours) & ${{224}^{2}}$   & 29M   & 5.1G     & \textbf{83.5}     \\
      \hline 
       ViT-B  \cite{DBLP:journals/corr/abs-2010-11929} & ${{384}^{2}}$   & 86M   & 55.4G     &71.6     \\
      Swin-B \cite{Liu-2021-ICCV} & ${{224}^{2}}$   & 88M   & 15.4G     &79.2     \\
      Focal-B \cite{DBLP:journals/corr/abs-2107-00641} & ${{224}^{2}}$   & 90M   & 16.0G     &79.1     \\
      CSWin-B \cite{Dong-2022-CVPR} & ${{224}^{2}}$   & 78M   & 15.0G     &82.5     \\
      MaxViT-B \cite{10.1007/978-3-031-20053-3-27} & ${{224}^{2}}$   & 120M   & 23.4G     &83.1     \\
 	 SGFormer-B \cite{Ren-2023-ICCV} & ${{224}^{2}}$   & 78M   & 15.6G     &82.7     \\
      VBB-B (ours) & ${{224}^{2}}$   & 85M   & 15.8G     &\textbf{83.7}     \\
      \hline 
   \end{tabular}
   }
   \label{Cifar-100-Top1}
\end{table}

\subsection{Classiﬁcation on Cifar-100 and Mini-ImageNet}

\noindent \textbf{Implementation details.} Follow the experimental settings in the above subsection.

\noindent \textbf{Results.} In Table \ref{Cifar-100-Top1} and Table \ref{Mini-ImageNet-Top1}, we compare the proposed VBB with state-of-the-art Transformer architectures on small datasets. With the limitation of pages, we only compare with a few classical methods here. It is known that ViTs usually perform poorly on such tasks as they typically require large datasets to be trained on. The models that perform well on large-scale ImageNet do not necessarily work perform on small-scale Mini-ImageNet and Cifar-100, e.g., ViT-B has top-1 accuracy of 75.7\% and Swin-B has top-1 accuracy of 82.3\% on the Mini-ImageNet, which suggests that ViTs are more challenging to train with less data. The proposed VBB can significantly improve the data efficiency and performs well on small datasets such as Cifar-100 and Mini-ImageNet. Compared with MaxViT-T, it has increased by 0.9\% and 2.1\% respectively.

\begin{table*}[t]
   \centering
   \caption{Object detection and instance segmentation performance on the COCO val2017 with the Mask R-CNN framework and 1x training schedule. The models have been pre-trained on ImageNet-1K. The resolution used to calculate FLOPs is 800×1280.}
   \begin{tabular}{l|cc|cccccc}
      \hline 
      Backbone  & Params & FLOPs & $\text{AP}^{\text{box}}$ &  $\text{AP}^{\text{box}}_{\text{50}}$  &  $\text{AP}^{\text{box}}_{\text{75}}$ &  $\text{AP}^{\text{mask}}$ &  $\text{AP}^{\text{mask}}_{\text{50}}$ &  $\text{AP}^{\text{mask}}_{\text{75}}$          \\
      \hline 
      ResNet-50 \cite{He-2016-CVPR} & 44M   & 260G   &38.0  &58.6 &41.4&34.4&55.1&36.7     \\
      Twins-S \cite{NEURIPS2021-4e0928de} & 44M   & 228G   &42.7&65.6 &46.7&39.6&62.5&42.6    \\
      PVT-S \cite{Wang-2021-ICCV} & 44M   & 245G   &40.4&62.9 &43.8&37.8&60.1&40.3     \\
      Swin-T \cite{Liu-2021-ICCV} & 48M   & 264G   &43.7&66.6 &47.6&39.8&63.3&42.7    \\
      Focal-T \cite{DBLP:journals/corr/abs-2107-00641} & 49M   & 291G   &44.8&67.7 &49.2&41.0&64.7&44.2    \\
      CSWin-T \cite{Dong-2022-CVPR} & 42M   & 279G   &46.7&68.6 &51.3&42.2&65.6&45.4     \\
	 SGFormer-S \cite{Ren-2023-ICCV} & 41M   & 275G   &47.4&69.0 &52.0&42.6&65.9&46.0     \\
      VBB-T (ours) & 45M   & 286G   &\textbf{47.7}&\textbf{69.3} &\textbf{52.1}&\textbf{42.6}&\textbf{66.3}&\textbf{46.3}          \\
      \hline 
      RegNeXt-101-64 \cite{He-2016-CVPR} & 101M   & 493G   &42.8  &63.8 &47.3&38.4&60.6&41.3     \\
      Twins-L \cite{NEURIPS2021-4e0928de} & 120M   & 474G   &45.2&67.5 &49.4&41.2&64.5&44.5    \\
      PVT-L \cite{Wang-2021-ICCV} & 81M   & 364G   &42.9&65.0 &46.6&39.5&61.9&42.5     \\
      Swin-B \cite{Liu-2021-ICCV} & 107M   & 496G   &46.9&-- &--&42.3&--&--      \\
      Focal-B \cite{DBLP:journals/corr/abs-2107-00641} & 110M   & 533G   &47.8&70.2 &52.5&43.2&67.3&46.5    \\
      CSWin-B \cite{Dong-2022-CVPR} & 97M   & 526G   &48.7&70.4 &53.9&43.9&67.8&47.3     \\
	 SGFormer-B \cite{Ren-2023-ICCV} & 95M   & 511G   &49.2&70.6 &54.3&44.1&68.1&47.7     \\
      VBB-B (ours) & 105M & 530G &\textbf{49.3}&\textbf{70.9} &\textbf{54.5}&\textbf{44.2}&\textbf{68.4}&\textbf{47.9} \\
      \hline 
   \end{tabular}
   \label{COCO-Top1}
\end{table*}

\subsection{Scaling Weight}

\begin{figure}[h]
  \centering
   \includegraphics[width=1.0\linewidth]{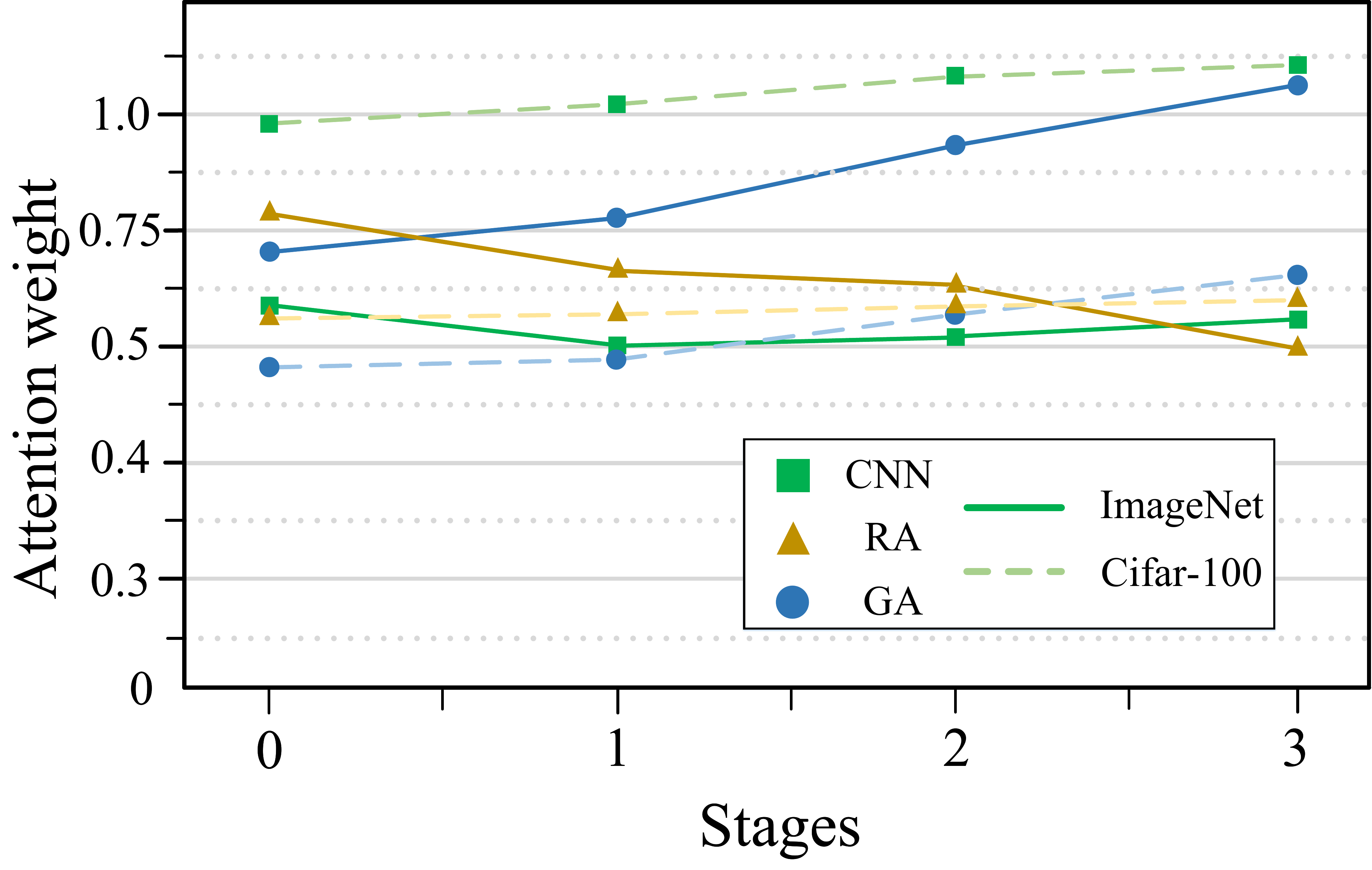}
   \caption{The average values of scaling weights at different stages of the VBB in ImageNet and Cifar-100 experiments.}
   \label{ScalingWeight-flabel}
\end{figure}

VBB adds scaling weights to the output of each mechanism, as shown in Figure \ref{OverallArchitecture-flabel} ($\alpha$,$\beta$,$\lambda$). We counted the average value of scaling weights in different stages of VBB in ImageNet and cifar100 experiments, as shown in Figure \ref{ScalingWeight-flabel}.

It can be seen from Figure \ref{ScalingWeight-flabel} that in the ImageNet, the scaling value of Global Attention in the shallow layer of the model is small and increases with the deepening of the layers, while that of RS-Win is opposite.  In addition, we find that the scaling values of CNNs in Cifar100 experiment is much larger than that in ImageNet, which to some extent can prove the effectiveness of CNN in feature extraction in small datasets.

\begin{table}[h]
   \centering
   \caption{Comparison of different models on Mini-ImageNet.}
   \resizebox{\linewidth}{!}{
   \begin{tabular}{l|ccc|c}
      \hline 
      Method  & Image Size & Param. & FLOPs                & Top-1 acc.           \\
      \hline 
      Swin-T \cite{Liu-2021-ICCV} & ${{224}^{2}}$   & 29M   & 4.5G     &82.1    \\
      Focal-T \cite{DBLP:journals/corr/abs-2107-00641} & ${{224}^{2}}$   & 29M   & 4.9G     &82.0     \\
      CSWin-T \cite{Dong-2022-CVPR} & ${{224}^{2}}$   & 23M   & 4.3G     &83.5     \\
      MaxViT-T \cite{10.1007/978-3-031-20053-3-27} & ${{224}^{2}}$   & 31M   & 5.6G     &84.1     \\
      SGFormer-S \cite{Ren-2023-ICCV} & ${{224}^{2}}$   & 22M   & 4.8G     &84.4     \\
      VBB-T (ours) & ${{224}^{2}}$   & 29M   & 5.1G     & \textbf{86.2}     \\
      \hline 
      ViT-B  \cite{DBLP:journals/corr/abs-2010-11929} & ${{384}^{2}}$   & 86M   & 55.4G     &75.7     \\
      Swin-B \cite{Liu-2021-ICCV} & ${{224}^{2}}$   & 88M   & 15.4G     &82.3     \\
      Focal-B \cite{DBLP:journals/corr/abs-2107-00641} & ${{224}^{2}}$   & 90M   & 16.0G     &82.5     \\
      CSWin-B \cite{Dong-2022-CVPR} & ${{224}^{2}}$   & 78M   & 15.0G     &83.8     \\
      MaxViT-B \cite{10.1007/978-3-031-20053-3-27} & ${{224}^{2}}$   & 120M   & 23.4G     &84.3     \\
 	 SGFormer-B \cite{Ren-2023-ICCV} & ${{224}^{2}}$   & 78M   & 15.6G     &84.6     \\
      VBB-B (ours) & ${{224}^{2}}$   & 85M   & 15.8G     &\textbf{86.5}     \\
      \hline 
   \end{tabular}
   }
   \label{Mini-ImageNet-Top1}
\end{table}

\subsection{COCO Object Detection}

\noindent \textbf{Implementation details.} We use the Mask R-CNN \cite{He-2017-ICCV} framework to evaluate the performance of the proposed VBB backbone on the COCO benchmark for object detection and instance segmentation. We pretrain the backbones on the ImageNet-1K dataset and apply the ﬁnetuning strategy used in Swin Transformer \cite{Liu-2021-ICCV} on the COCO training set.

\noindent \textbf{Results.} We compare VBB with various backbones, as shown in Table \ref{COCO-Top1}. It shows that the proposed VBB variants clearly outperform all the CNN and Transformer counterparts. For object detection, our VBB-T and VBB-B achieve 47.7 and 49.3 box mAP for object detection, surpassing the previous best SGFormer by +0.3 and +0.1, respectively. We also achieve similar performance gain on instance segmentation. Such supremacy comes from the ability to preserve more fine-granularity information and the global receptive field in VBB.

\begin{table}[t]
   \centering
   \caption{Comparison of the segmentation performance of different backbones on the ADE20K. All backbones are pretrained on ImageNet-1K with the size of 224 ×224. The resolution used to calculate FLOPs is 512 ×2048.}
   \resizebox{\linewidth}{!}{
   \begin{tabular}{l|cc|cc}
      \hline 
      Backbone  & Params & FLOPs & SS mIoU & MS mIoU     \\
      \hline 
      Twins-S \cite{NEURIPS2021-4e0928de} & 55M   & 905G   &46.2&47.1     \\
      Swin-T \cite{Liu-2021-ICCV} & 60M   & 945G   &44.5&45.8    \\
      Focal-T \cite{DBLP:journals/corr/abs-2107-00641} & 62M   & 998G   &45.8&47.0    \\
      CSWin-T \cite{Dong-2022-CVPR} & 60M   & 959G   &49.3&50.7     \\
	 SGFormer-S \cite{Ren-2023-ICCV} & 53M   & 989G   &49.9&51.5     \\
      VBB-T (ours) & 57M   & 996G   &\textbf{50.3}&\textbf{51.8}           \\
      \hline 
      Twins-L \cite{NEURIPS2021-4e0928de} & 113M   & 1164G   &48.8&50.2     \\
      Swin-B \cite{Liu-2021-ICCV} & 121M   & 1188G   &48.1&49.7     \\
      Focal-B \cite{DBLP:journals/corr/abs-2107-00641} & 126M   & 1354G   &49.0&50.5     \\
      CSWin-B \cite{Dong-2022-CVPR} & 109M   & 1222G   &51.1&52.2     \\
	 SGFormer-B \cite{Ren-2023-ICCV} & 109M   & 1304G   &52.0&52.7     \\
      VBB-B (ours) & 115M & 1328G &\textbf{52.5}&\textbf{52.9}  \\
      \hline 
   \end{tabular}
   \label{ADE20K-Top1}
   }
\end{table}

\subsection{ADE20K Semantic Segmentation}

\noindent \textbf{Implementation details.} We further investigate the capability of VBB for Semantic Segmentation on the ADE20K \cite{Zhou-2017-CVPR} dataset. Here we employ the widely-used UperNet \cite{Xiao-2018-ECCV} as the basic framework and followed Swin's \cite{Liu-2021-ICCV} experimental settings In Table \ref{ADE20K-Top1}, we report both the single-scale (SS) and multi-scale (MS) mIoU for better comparison.

\noindent \textbf{Results.} As shown in Table \ref{ADE20K-Top1}, our VBB variants outperform previous state-of-the-arts under different conﬁgurations. Speciﬁcally, our VBB-T and VBB-B outperform the SGFormer by +0.4\% and +0.5\% SS mIoU, respectively. These results show that the proposed VBB can effectively capture the context dependencies of different distances.

\begin{table}[h]
   \centering
   \caption{Comparing the impact of each mechanism on VBB performance.}
   \begin{tabular}{l|ccc}
      \hline 
        & ImageNet & COCO & ADE20k                    \\
         & top-1   & $\text{AP}^{\text{box}}$    &SS mIoU     \\
      \hline 
	  Remove \textit{CNN}  & 82.5    &47.2  &49.7    \\
       Remove \textit{RS-Win}  & 83.1    &46.8  &49.3    \\
       Remove \textit{GA}  & 83.4    &46.5  &49.1    \\
       VBB   & \textbf{83.8}    &\textbf{47.7}  &\textbf{50.3}    \\
      \hline 
   \end{tabular}
   \label{CombinedUse}
\end{table}

\subsection{Ablation Study}

We perform ablation studies on image classification and downstream tasks for the fundamental designs of our VBB. For a fair comparison, we only change one component for each ablation.

\noindent \textbf{Different mechanisms.} In this subsection, we show the impact of removing one feature extraction mechanism from the VBB on the performance of the VBB. As shown in Table \ref{CombinedUse}, the combination works best. It can also be seen that \textit{CNN} is more important for recognition tasks, while \textit{GA} is more important for object detection and semantic segmentation.

\begin{table}[h]
   \centering
   \caption{Comparing VBB with and without positional encoding. \textbf{APE}: absolute positional encoding. \textbf{RPE}: relative positional encoding.}
   \begin{tabular}{l|ccc}
      \hline 
        & ImageNet & COCO & ADE20k                    \\
         & top-1   & $\text{AP}^{\text{box}}$    &SS mIoU     \\
      \hline 
	  add APE  & 83.5    &46.9  &50.0    \\
       add RPE  & \textbf{83.9}    &47.5  &50.2    \\
       without PE   & {83.8}    &\textbf{47.7}  &\textbf{50.3}    \\
      \hline 
   \end{tabular}
   \label{VBBwithoutPE}
\end{table}

\noindent \textbf{VBB without Positional Encoding.} In this subsection, we compare VBB with and without positional encoding. As shown in Table \ref{VBBwithoutPE}, removing positional encoding in the VBB does not have a significant impact on performance.

\begin{table}[h]
   \centering
   \caption{Comparison of  different self-attention  mechanisms.}
   \resizebox{\linewidth}{!}{
   \begin{tabular}{l|ccc}
      \hline 
        & ImageNet & COCO & ADE20k                    \\
         & top-1   & $\text{AP}^{\text{box}}$    &SS mIoU     \\
      \hline 
       Swin's shifted windows \cite{Liu-2021-ICCV}   & 82.3    &45.7  &48.5    \\
       Sequential Axial \cite{DBLP:journals/corr/abs-1912-12180}   & 82.5    &45.5  &48.9    \\
       Cross-shaped  \cite{Dong-2022-CVPR}   & 82.7    &46.7  &49.3    \\
       Grid \cite{10.1007/978-3-031-20053-3-27}   & 83.6    &46.9  &49.5    \\
	  SG \cite{Ren-2023-ICCV}   & 83.2    &47.4  &49.9    \\
       VBB (ours)   & \textbf{83.8}    &\textbf{47.7}  &\textbf{50.3}    \\
      \hline 
   \end{tabular}
   }
   \label{differentMechanisms}
\end{table}

\noindent \textbf{Attention Mechanism Comparison.} In this subsection, we compare with existing self-attention mechanisms. As shown in Table \ref{differentMechanisms}, the proposed VBB self-attention mechanism performs better than the existing self-attention mechanism.

\section{Conclusions}

In this paper, we propose a new Vision Transformer architecture named VBB. The core design of VBB consists of three components: \textit{Convolution}, Random Sampling Window (\textit{RS-Win}) and Global Attention (\textit{GA}). \textit{Convolution} captures the fine-grained local information. \textit{RS-Win} can be seen as the result of sparsification of full attention. \textit{GA} supplements the global information for VBB. On the other hand, VBB performs \textit{Convolution}, \textit{RS-Win} and \textit{GA} in parallel by splitting the multiheads into three parallel groups. This multi-head grouping design allows the model to efficiently incorporate information from the three components without extra computation cost. VBB is a universal approximator of sequence functions and is Turing complete, which can achieve state-of-the-art performance on ImageNet-1K image classification, COCO object detection and ADE20K semantic segmentation.

\bibliographystyle{named}
\bibliography{ijcai22}

\end{document}